# An Efficient Implementation of Belief Function Propagation

**Hong Xu**
IRIDIA - Université Libre de Bruxelles
50 av. F. Roosevelt, CP 194/6
B-1050, Brussels, Belgium
Email: r01505@bbrbfu01.bitnet

## Abstract

The local computation technique (Shafer et al. 1987, Shafer and Shenoy 1988, Shenoy and Shafer 1986) is used for propagating belief functions in so-called a Markov Tree. In this paper, we describe an efficient implementation of belief function propagation on the basis of the local computation technique. The presented method avoids all the redundant computations in the propagation process, and so makes the computational complexity decrease with respect to other existing implementations (Hsia and Shenoy 1989, Zarley et al. 1988). We also give a combined algorithm for both propagation and re-propagation which makes the re-propagation process more efficient when one or more of the prior belief functions is changed.

## 1 INTRODUCTION

Dempster-Shafer theory (Shafer 1976, Smets 1988) has been considered as one of the tools for dealing with the problem of uncertain information by the Artificial Intelligence community. The computational complexity of Dempster's rule of combination, the pivot mechanism of the theory, however, is the main obstacle to its effective use. However, several implementations of the Dempster-Shafer theory have recently been developed (Hsia and Shenoy 1989, Zarley 1988, Zarley et al. 1988) based on the observation that an arbitrary belief function network can be represented as a hypergraph (Kong 1986), which can also be embedded in so-called a Markov tree (Zhang 1988). These implementations use the local computation technique (Shafer et al. 1987, Shafer and Shenoy 1988, Shenoy and Shafer 1986) for propagating belief functions in the Markov tree. According to this technique, the belief function propagation can be described as a message-passing scheme: each node in the Markov tree sends its message to one of its neighbours after it has received the messages from all of its other neighbours, and the result of propagation on each node is computed by combining its own belief function (prior belief function) and the messages from all of its neighbours. After the results for all the nodes have been computed, one may want to change one or more of the prior belief functions. Then we have to re-propagate the impact of the changes to all the other nodes. In general, there may be repeated computations during propagation and re-propagation process, which may greatly affect the efficiency of the computation. The goal of this paper is to present an efficient method for the implementation of belief function propagation. The main advantage of this scheme is that it avoids all the redundant computations during propagation, resulting in a reduced computational complexity with respect to that of other existing implementations. It is also shown that making full use of the stored messages passed between the nodes and of stored intermediate information, we can just re-propagate the changed values when some prior belief functions are changed.

The paper is organized as follows. In section 2, some basic concepts about belief function networks are reviewed. In section 3, we describe a straightforward implementation of belief functions propagation using local computation. In section 4, we present our implementation scheme. In section 5, we discuss the problem of updating messages when one or more inputs is changed, and give a combined algorithm for both propagation and re-propagation. Finally, some conclusions are given in section 6.

## 2 SOME BASIC CONCEPTS ABOUT BELIEF FUNCTION NETWORKS

Dempster-Shafer theory (Shafer 1976, Smets 1988), is concerned with the problem of representing and manipulating incomplete knowledge. In this section, we recall some basic concepts and definitions about belief functions and belief function networks. This presentation follows (Shafer and Shenoy 1988, Shenoy 1989).

<u>Variables and Configurations</u> We use the symbol $\mathcal{W}_X$ for the set of possible values of a variable X, and we call $\mathcal{W}_X$ the *frame for X.* Given a finite non-empty set h of variables, let $\mathcal{W}_h$ denote the Cartesian product of $\mathcal{W}_X$ for X in h: $\mathcal{W}_h = \times\{\mathcal{W}_X \mid X \in h\}$. We call $\mathcal{W}_h$ the *frame for h.* We refer to elements of $\mathcal{W}_h$ as *configurations of h.*

<u>Basic Probability Assignments</u> A *basic probability assignment (bpa)* m on X, is a function which assigns a value in [0, 1] to every subset $\mathfrak{a}$ of $\mathcal{W}_X$ and satisfies the following axioms:
(i) $m(\emptyset) = 0$; and
(ii) $\Sigma\{m(\mathfrak{a}) \mid \mathfrak{a} \subseteq \mathcal{W}_X\} = 1$

**Belief Functions** A *belief function* Bel associated with a bpa m, is a function that assigns a value in [0, 1] to every non-empty subset $\mathfrak{a}$ of $\mathcal{W}_X$, called "degree of belief in $\mathfrak{a}$", defined by

$$\mathrm{Bel}(\mathfrak{a}) = \Sigma\{m(\mathfrak{b}) \mid \mathfrak{b} \subseteq \mathfrak{a}\}$$

The subsets $\mathfrak{a}$ for which $m(\mathfrak{a})>0$ are called *focal elements* of Bel. The simplest belief function is the one with $m(\mathcal{W}_X) = 1$, called vacuous belief function.

**Projection and Extension** If g and h are sets of variables, $h \subseteq g$, and x is a configuration of g, then we let $x^{\downarrow h}$ denote the projection of x to $\mathcal{W}_h$. $x^{\downarrow h}$ is always a configuration of h. If $\mathfrak{g}$ is a non-empty subset of $\mathcal{W}_g$, then the *projection of* $\mathfrak{g}$ *to h*, denoted by $\mathfrak{g}^{\downarrow h}$, is obtained by projecting elements of $\mathfrak{g}$ to $\mathcal{W}_h$, i.e. $\mathfrak{g}^{\downarrow h} = \{x^{\downarrow h} \mid x \in \mathfrak{g}\}$. By extension of a subset of a frame to a subset of a larger frame, we mean a cylinder set extension. If g and h are sets of variables, $h \subseteq g$, $h \neq g$, and $\mathfrak{h}$ is a subset of $\mathcal{W}_h$, then the *extension of* $\mathfrak{h}$ *to g*, denoted by $\mathfrak{h}^{\uparrow g}$, is $\mathfrak{h} \times \mathcal{W}_{g-h}$.

**Dempster's Rule of Combination** *Dempster's Rule of Combination* is a rule for producing a new bpa from two bpa's. Considering two bpa's $m_1$ and $m_2$ on g and h, we let $m = m_1 \oplus m_2$ be the bpa on $g \cup h$ defined by

$$m(\emptyset) = 0 \text{ and}$$

$$m(\mathfrak{c}) = K^{-1}\Sigma\{m_1(\mathfrak{a})m_2(\mathfrak{b}) \mid (\mathfrak{a}^{\uparrow(g \cup h)} \cap \mathfrak{b}^{\uparrow(g \cup h)}) = \mathfrak{c}\}$$

where $K = 1 - \Sigma\{m_1(\mathfrak{a})\, m_2(\mathfrak{b}) \mid (\mathfrak{a}^{\uparrow(g \cup h)} \cap \mathfrak{b}^{\uparrow(g \cup h)}) = \emptyset\}$

K is a normalizing factor, which intuitively measures how much $m_1$ and $m_2$ are conflicting. If K = 0, then we say that $m_1$ and $m_2$ are not combinable.

**Marginalization** Suppose m is a bpa on g and suppose $h \subseteq g$, $h \neq \emptyset$. *The marginal of m for h*, denoted by $m_{\downarrow h}$, is a bpa defined by

$$m^{\downarrow h}(\mathfrak{a}) = \Sigma\{m(\mathfrak{b}) \mathfrak{b} \subseteq \mathcal{W}_g, \mathfrak{b}^{\downarrow h} = \mathfrak{a}\} \text{ for all subsets } \mathfrak{a} \text{ of } \mathcal{W}_h.$$

**A Belief Function Network** Using Dempster-Shafer theory, the problem can be represented as a finite set of variables $\mathcal{X}$. A finite frame $\mathcal{W}_X$ is associated with each variable X of $\mathcal{X}$ and the elements of $\mathcal{W}_X$ are mutually exclusive and exhaustive.The relationships of the variables are expressed by subsets of $\mathcal{X}$. The knowledge about the variable X is encoded in the belief function over $\mathcal{W}_X$ or over $\mathcal{W}_h$ where h = {X}. The knowledge about relation among the variables is encoded in the belief function over $\mathcal{W}_h$ of subset h in w hich the variables are included. We call these belief functions as prior belief functions. So, *a belief function network* consists of $\mathcal{X}$, a set of subsets of $\mathcal{X}$: $\mathcal{H}$, and a finite collection of independent belief functions ($\mathrm{Bel}_1$, $\mathrm{Bel}_2$, ..., $\mathrm{Bel}_k$) where each belief function $\mathrm{Bel}_i$ is the prior belief function on some subset h, and is stored as a set of focal elements with their values.

**Evaluation of a Belief Function Network** Suppose we are given a belief function network. To evaluate a belief function network, we have to:
(i) combine all $\mathrm{Bel}_i$ in the network, the resulting belief function is called *global belief function* ;
(ii) compute the marginal of the global belief function for each variable in the network.

Because the computational complexity of Dempster's combination is exponential with the size of the frame of belief functions being combined, it will not be feasible to compute the global belief function when there are a large number of variables. In the next section, we will describe an alternative way to evaluate the belief function network by using the local computation technique proposed in (Shafer et al. 1987, Shafer and Shenoy 1988, Shenoy and Shafer 1986).

## 3 BELIEF FUNCTION PROPAGATION USING LOCAL COMPUTATION

It has been shown that if the belief function network can be represented as certain kind of tree, called Markov tree, the belief functions can be "propagated" in the Markov tree by a local message-passing scheme, producing as a result in the marginals of the global belief function for each of the nodes. We first look at what Markov tree is and how a belief function network can be represented as a Markov tree.

Given a tree $G=(\mathcal{N}, \mathcal{E})$ where each node (also called vertex) $v \in \mathcal{N}$ is a non-empty subset of a finite set V, $\mathcal{E}$ is the set of edges in G. Then G is *Markov* if for any $v \in \mathcal{N}$, such that v separates two other distinct nodes $v_i$ and $v_j$ in G, $(v_i \cap v_j) \subseteq v$. Given three distinct nodes v, $v_i$ and $v_j$, we say that v *separates* $v_i$ and $v_j$ if v is on the path between $v_i$ and $v_j$.

Let $\mathcal{H}$ and $\mathcal{X}$ be as defined in the previous section. In the language of graph theory, $\mathcal{H}$ is called a *hypergraph* on $\mathcal{X}$ and each element of $\mathcal{H}$ is called a *hyperedge*. In order to use local computation for propagation, the hypergraph should be arranged in a Markov tree where $V=\mathcal{X}$ and $\mathcal{N} \supseteq \mathcal{H}$. We can always find a method to arrange a hypergraph in a Markov tree. Algorithms for constructing a Markov tree for a hypergraph can be found in (Kong 1986, Mellouli 1987, Zhang 1988). Two examples of Markov tree representatives (on the right hand side of Fig 3.1) for hypergraphs (on the left hand side of Fig 3.1) are shown below. In Example1, $\mathcal{X}_1 = \{a, b, c\}$, $\mathcal{H}_1 = \{\{a\}, \{b\}, \{c\}, \{a, b\}, \{b, c\}\}$, $\mathcal{N}_1 = \mathcal{H}_1$; In Example2, $\mathcal{X}_2 = \{p, q, s, t, r\}$, $\mathcal{H}_2 = \{\{s\}, \{t\}, \{p\}, \{q\}, \{r\}, \{s, p\}, \{p, t\}, \{t, q\}, \{s, q\}, \{p, r\}\}$, where $\mathcal{N}_2 = \mathcal{H}_2 \cup \{\{p, q, t\}, \{s, p, q\}\}$, where {p, q, t} and {s, p, q} are the new nodes added for constructing the Markov tree.

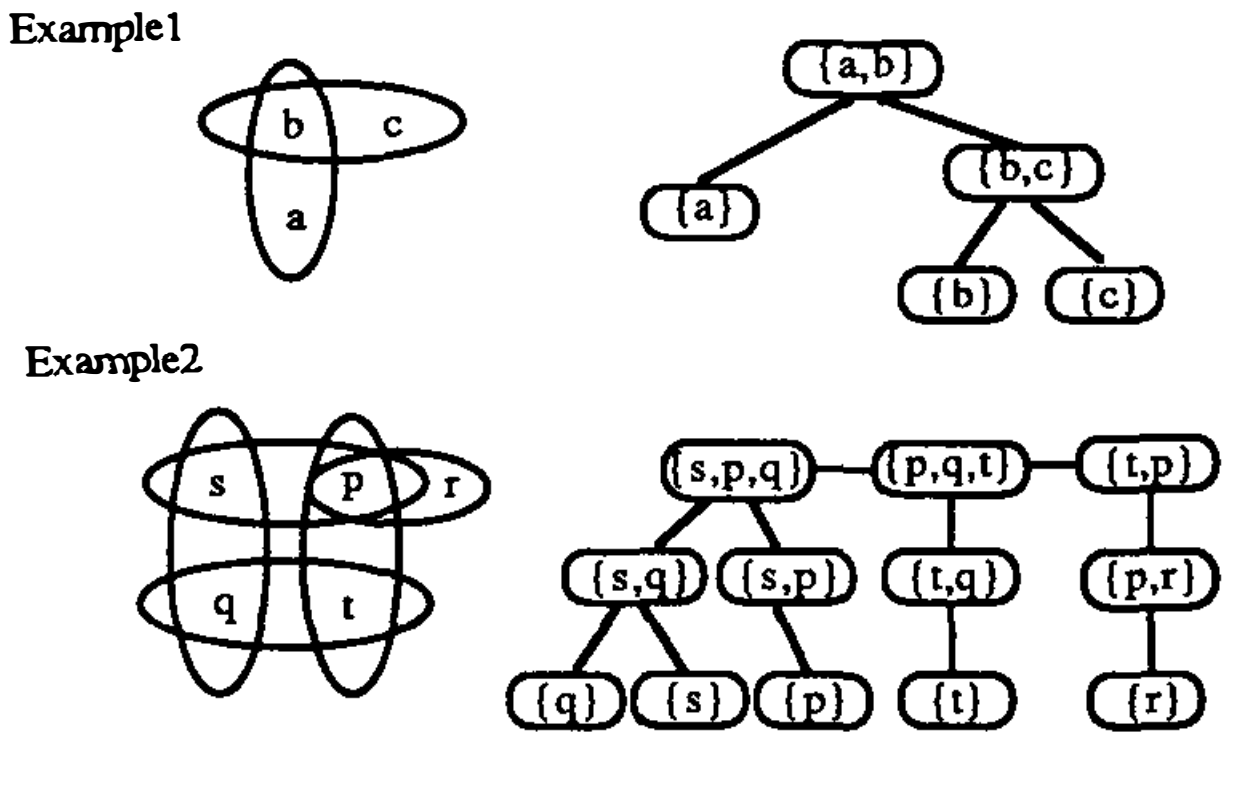

Fig3.1: Markov tree representatives of hypergraphs

In the rest of this section, we discuss Shafer, Shenoy and Mellouli's propagation scheme using local computation (Shafer et al. 1987). Suppose we have already arranged the hypergraph in a Markov tree G=($\mathcal{M}$, $\mathcal{E}$). For each node v, we let $\mathcal{N}_v=\{v_k|(v_k, v)\in \mathcal{E}\}$ be the set of neighbours of v, $Bel_v$ the prior belief function on v, and $Bel^{\downarrow v}$ the marginal of the global belief function for v. Let L(G) be the leaves of G given some designated node as the root of G. During propagation, each node sends a belief function to each of its neighbours. The belief function sent by v to $v_i$ is referred as a "message" and is denoted by $M^{v\to v_i}$. We define it as:

$$M^{v\to v_i}=((Bel_v\oplus(\oplus \{M^{v_k\to v}|v_k\in(\mathcal{N}_v-\{v_i\})\}))^{\downarrow(v\cap v_i)})^{\uparrow v_i} \qquad (3.1)$$

Because a leaf v has only one neighbour, say $v_i$, then the above expression reduces to:

$$M^{v\to v_i} = ((Bel_v)^{\downarrow(v\cap v_i)})^{\uparrow v_i}$$

Thus, when the propagation starts, the leaves of the Markov tree can send messages to their neighbours right away. The others send a message to one neighbour after they have received messages by all but that one neighbour. And when a node receives a message from that one neighbour, it appropriately (i.e. by using (3.1)) sends messages back to the remaining neighbours. All the messages can be transmitted through the Markov tree in this way.

After node v has received the messages from all the neighbours, the marginal $Bel^{\downarrow v}$ for v is given by

$$Bel^{\downarrow v}=Bel_v\oplus(\oplus\{M^{v_i\to v} \mid v_i\in \mathcal{N}_v\}) \qquad (3.2)$$

Because all the variables are included in the Markov tree, as Fig 3.1 illustrated, we can simultaneously compute the marginals of the global belief function for every variables in the belief function network. The whole propagation process is shown in Fig 3.2. For more detail about this propagation scheme, see (Saffiotti 1989, Shafer et al. 1987, Shafer and Shenoy 1988, Shenoy and Shafer 1986).

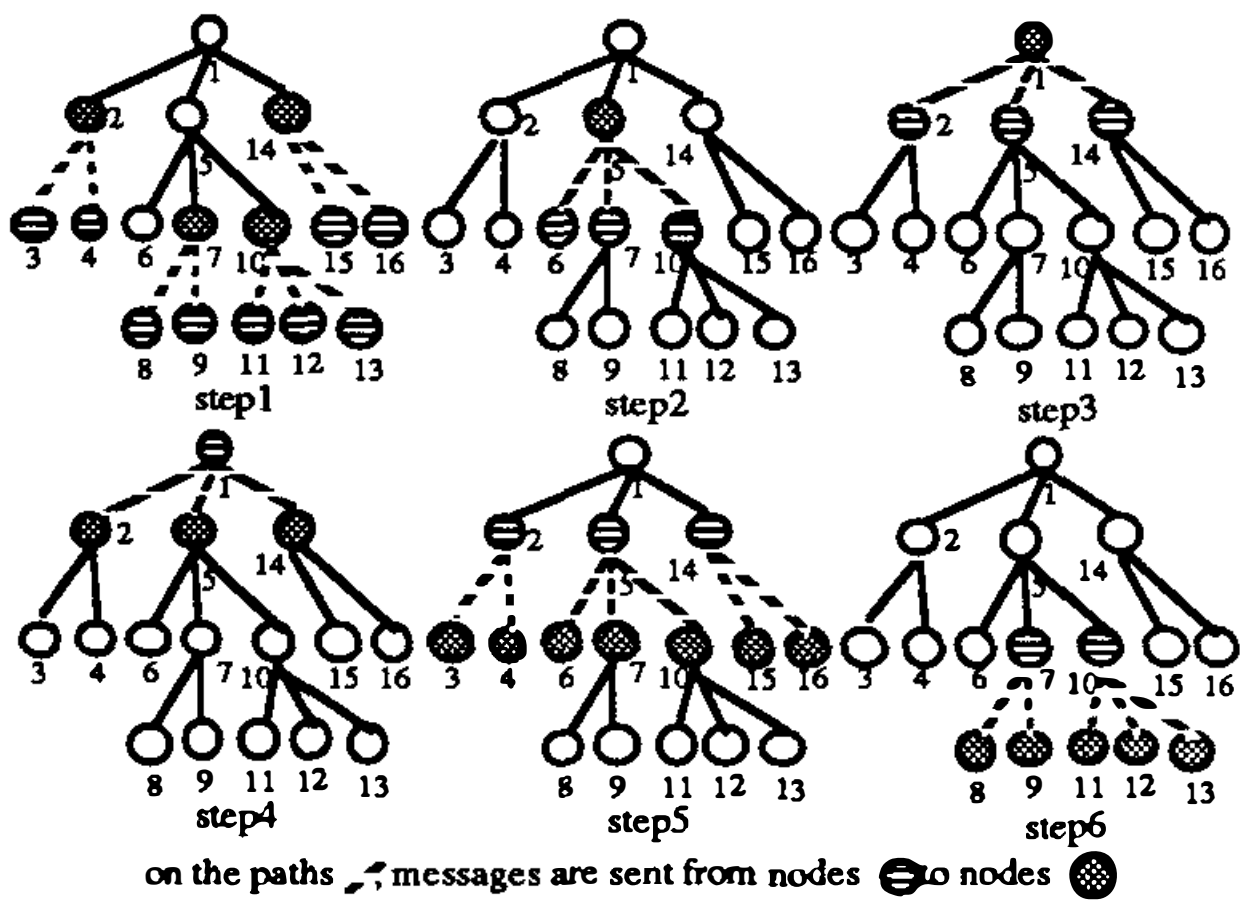

Fig3.2: the message-passing scheme for simultaneous belief function propagation

According to the scheme described above, a typical message-passing situation can be illustrated as in Fig 3.3. A straightforward way for computing the messages between the nodes and the marginals for the nodes is as follows.

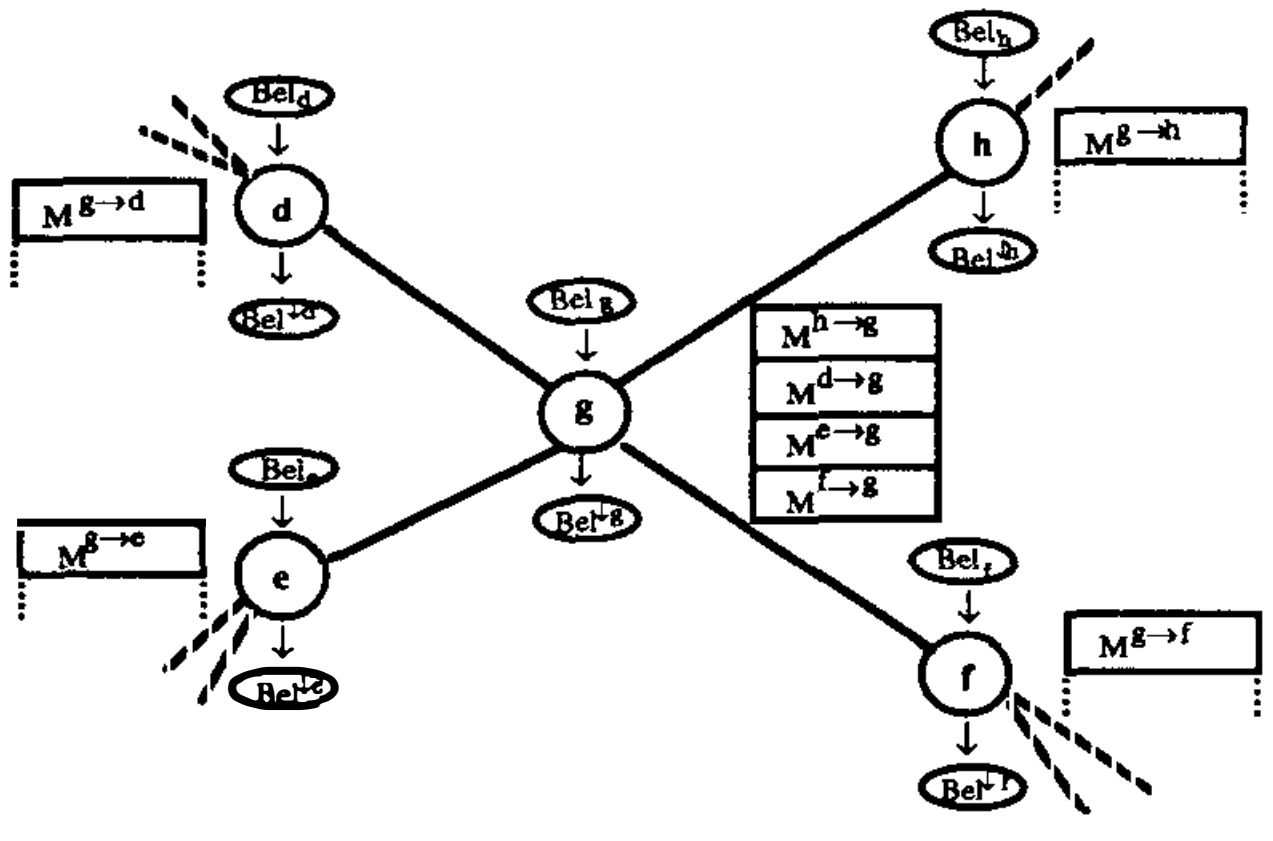

Fig3.3: a typical message-passing situation

<u>Example 3.1:</u> Suppose nodes d, e and f have received messages from all their respective neighbours except node g. Suppose now we want to compute the marginals for d, e and f. According to (3.1), we first compute all $M^{i\to g}$, $i\in\{d, e, f\}$

$$M^{i\to g}=((Bel_i\oplus\{\oplus M^{z\to i}|z\in \mathcal{N}_i-\{g\}\})^{\downarrow(i\cap g)})^{\uparrow g} \qquad (3.3)$$

Now g has received the messages by d, e and f, so it can send message to h. Again using (3.1), we have:

$$M^{g\to h}=((Bel_g\oplus M^{d\to g}\oplus M^{e\to g}\oplus M^{f\to g})^{\downarrow(g\cap h)})^{\uparrow h} \qquad (3.4)$$

Later, after g has received the message from h, it can send messages back to d, e and f, and the marginals for d, e and f will be computed. Using (3.2), we have

$$Bel^{\downarrow f}=Bel_f\oplus\{\oplus M^{z\to f}|z\in \mathcal{N}_f-\{g\}\}\oplus ((Bel_g\oplus M^{d\to g}\oplus M^{e\to g}\oplus M^{h\to g})^{\downarrow(g\cap f)})^{\uparrow f} \qquad (3.5)$$

$$Bel^{\downarrow e}=Bel_e\oplus\{\oplus M^{z\to e}|z\in\mathcal{N}_e-\{g\}\}\oplus((Bel_g\oplus M^{d\to g}\oplus M^{f\to g}\oplus M^{h\to g})^{\downarrow(g\cap e)})^{\uparrow e} \quad (3.6)$$

$$Bel^{\downarrow d}=Bel_d\oplus\{\oplus M^{z\to d}|z\in\mathcal{N}_d-\{g\}\}\oplus((Bel_g\oplus M^{e\to g}\oplus M^{f\to g}\oplus M^{h\to g})^{\downarrow(g\cap d)})^{\uparrow d} \quad (3.7)$$

Some repeated computations are found here. e.g. $Bel_f\oplus\{\oplus M^{z\to f}|z\in\mathcal{N}_f-\{g\}\}$ is computed four times in (3.3),(3.5), (3.6) and (3.7). $Bel_g\oplus M^{f\to g}\oplus M^{h\to g}$ is computed twice in (3.6) and (3.7). The solution will be discussed in the next section.

## 4 A More Efficient Implementation

It is well known that the computation of Dempster's combination involves the most computational expense during the whole propagation process. So the redundant combinations during the propagation should be avoided as much as possible. In this section, we present an algorithm for belief function propagation using local computation which avoids those repeated computations described above.

In our implementation, we assume that once we have chosen a Markov tree representative $G = (\mathcal{M},\mathcal{E})$, G will not change unless the belief function network which it represents is changed. We still make use of the notations defined in the preceding section. We choose one node of G, say $v_r$, to be the root of the tree, thus the edges in G can be seen as direct edges: we say that an edge $(v, v_i)$ in G is directed from v to $v_i$ whenever node $v_i$ is on the path between node v and $v_r$. In other words, we can define the parent-children relationship for each node v: let $\mathcal{C}h_v=\{c_k \mid c_k\in\mathcal{N}_v$, v is on the path between $c_k$ and $v_r\}$ be the children of v and $\mathcal{P}_v$ be the parent of v if $\mathcal{P}_v\in\mathcal{N}_v$ and $\mathcal{P}_v$ is on the path between v and $v_r$. We also assume that there is an order (arbitrary but fixed) for the elements in $\mathcal{C}h_v$, and let $\mathcal{C}h'_v$ denote the same set as $\mathcal{C}h_v$ but with reverse order. For each $c_k\in\mathcal{C}h_v$, we let $\mathcal{L}sb_{c_k}=\{c_j \mid c_j\in\mathcal{C}h_v, j<k\}$ denote the left hand siblings of $c_k$, and $\mathcal{R}sb_{c_k}=\{c_j \mid c_j\in\mathcal{C}h_v, j>k\}$ denote the right hand siblings of $c_k$. Furthermore, to remove redundant computations, we associate three intermediate variables $\mathcal{C}ur_v$, $\mathcal{I}ntm_v$ and $R_v$ with each node v. Suppose that for a given node v, it is $\mathcal{C}h_v=\{c_1, c_2, ..., c_m\}$. Then we give below the formulas for computing these intermediate variables:

$$\mathcal{C}ur_v=Bel_v\oplus\{\oplus M^{c_k\to v}|c_k\in\mathcal{C}h_v\} \quad (4.1)$$

$$\mathcal{I}ntm_{c_i}=Bel_v\oplus\{\oplus M^{c_k\to v}|c_k\in\mathcal{L}sb_{c_i}\} \quad (4.2)$$

$$R_{c_i}=M^{\mathcal{P}_v\to v}\oplus\{\oplus M^{c_k\to v}|c_k\in\mathcal{R}sb_{c_i}\}$$

Then we compute the marginal for $c_i$, one of the children of v, as follows:

$$\begin{aligned}Bel^{\downarrow c_i}&=\mathcal{C}ur_{c_i}\oplus((\mathcal{I}ntm_{c_i}\oplus R_{c_i})^{\downarrow(v\cap c_i)})^{\uparrow c_i}\\
&=Bel_{c_i}\oplus\{\oplus M^{z\to c_i}|z\in\mathcal{C}h_{c_i}\}\oplus\\
&\quad((Bel_v\oplus\{\oplus M^{c_k\to v}|c_k\in\mathcal{L}sb_{c_i}\}\oplus M^{\mathcal{P}_v\to v}\oplus\\
&\quad\{\oplus M^{c_k\to v}|c_k\in\mathcal{R}sb_{c_i}\})^{\downarrow(v\cap c_i)})^{\uparrow c_i}\\
&= Bel_{c_i}\oplus\{\oplus M^{z\to c_i}|z\in\mathcal{C}h_{c_i}\}\oplus((Bel_v\oplus\\
&\quad\{\oplus M^{c_k\to v}|c_k\in\mathcal{N}_v-\{c_i\}\})^{\downarrow(v\cap c_i)})^{\uparrow c_i}\\
&=Bel_{c_i}\oplus\{\oplus M^{z\to c_i}|z\in\mathcal{C}h_{c_i}\}\oplus\{M^{z\to c_i}|z=\mathcal{P}_{c_i}\}\\
&=Bel_{c_i}\oplus\{\oplus M^{v_j\to c_i}|v_j\in\mathcal{N}_{c_i}\}\end{aligned}$$

which is what (3.2) requires to be the case. For the root $v_r$, because $\mathcal{C}h_{v_r}=\mathcal{N}_{v_r}$, so, when node $v_r$ receives the messages from all of its children, the marginal for $v_r$ can be computed immediately, i.e. $Bel^{\downarrow v_r}=\mathcal{C}ur_{v_r}$.

From the analysis in Example 4.1, we will see how the number of applications of Dempster's combination can be reduced to the least. The typical message-passing situation shown in Fig 3.3 above is now as in Fig 4.1. The arrows in the edges are the directions of edges, and two more storages, $\mathcal{C}ur_i$ and $\mathcal{I}ntm_i$, are required at each node. Because $R_i$ is used just once, we do not store it at each node, but regard it as a temporary variable.

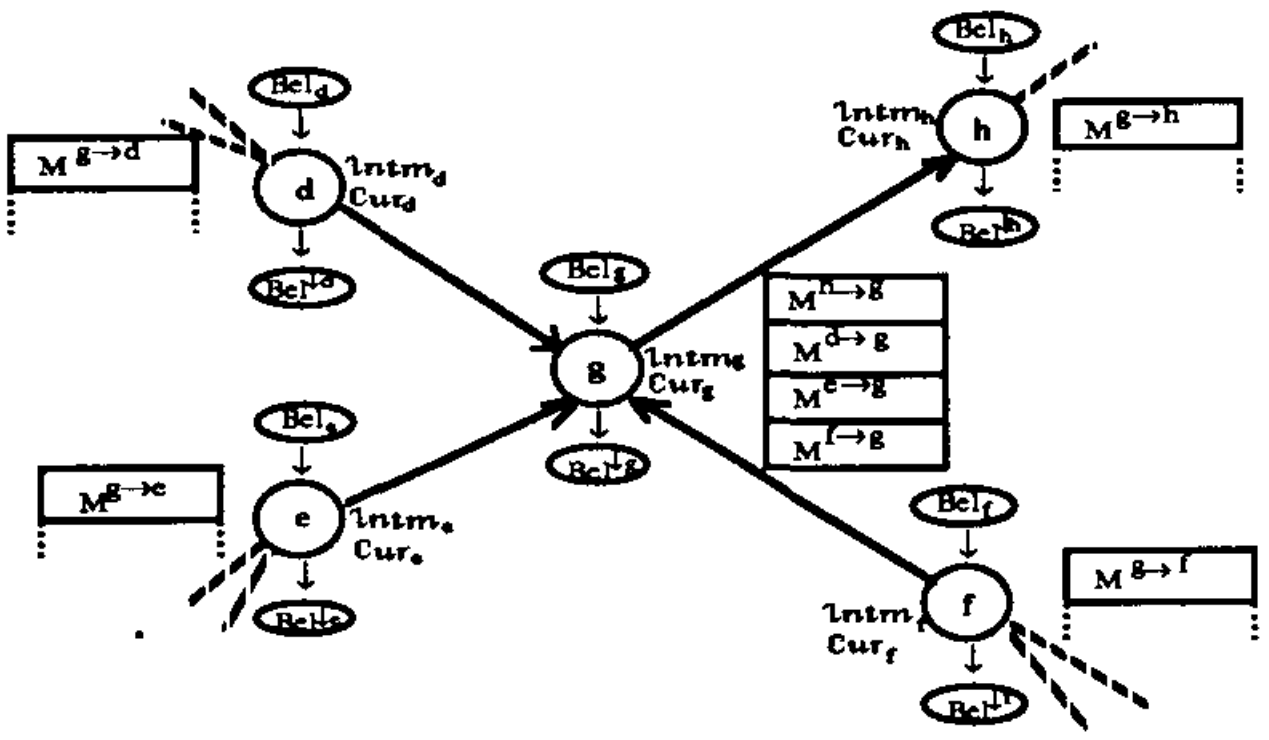

Fig4.1: a typical message-passing situation

Example 4.1: In this example, we consider the same situation as in Example 3.1. Because nodes d, e and f have received all the messages from their children, $\mathcal{C}ur_d$, $\mathcal{C}ur_e$ and $\mathcal{C}ur_f$ can be computed according to (4.1). Then the messages $M^{i\to g}$, i = d, e, f, can be computed as follows:

$$M^{i\to g}=((\mathcal{C}ur_i)^{\downarrow(i\cap g)})^{\uparrow g}$$

Now g can receive the messages from its children d, e and f in sequence:

$$\begin{aligned}\mathcal{I}ntm_d&=Bel_g\\
\mathcal{I}ntm_e&=\mathcal{I}ntm_d\oplus M^{d\to g}\\
\mathcal{I}ntm_f&=\mathcal{I}ntm_e\oplus M^{e\to g}\\
\mathcal{C}ur_g&=\mathcal{I}ntm_f\oplus M^{f\to g}\end{aligned}$$

$M^{g \to h} = ((\mathcal{Cur}_g)^{\downarrow(g \cap h)})^{\uparrow h}$

After h has sent the message back to g, g can send the messages back to its children f, e and d, and the marginals for d, e and f can be computed.

$R_f = M^{h \to g}$

$M^{g \to f} = ((\mathcal{Intm}_f \oplus R_f)^{\downarrow(g \cap f)})^{\uparrow f}$

$Bel^{\downarrow f} = \mathcal{Cur}_f \oplus M^{g \to f}$

$R_e = R_f \oplus M^{f \to g}$

$M^{g \to e} = ((\mathcal{Intm}_e \oplus R_e)^{\downarrow(g \cap e)})^{\uparrow e}$

$Bel^{\downarrow e} = \mathcal{Cur}_e \oplus M^{g \to e}$

$R_d = R_e \oplus M^{e \to g}$

$M^{g \to d} = ((\mathcal{Intm}_d \oplus R_d)^{\downarrow(g \cap d)})^{\uparrow d}$

$Bel^{\downarrow d} = \mathcal{Cur}_d \oplus M^{g \to d}$

Thus, all the repeated computation in (3.5), (3.6) and (3.7) of Example 3.1 are avoided by using $\mathcal{Cur}_i$, $\mathcal{Intm}_i$ and $R_i$. Table 4.1 illustrates the situation (only the number of combinations is compared). In this sense, our approach is an optimal implementation of propagation using local computation.

Table4.1 Comparison of the number of combinations used by our approach and the straightforward one

| to compute | Example 3.1 | Example 4.1 | comparison |
|---|---|---|---|
| $M^{g \to h}$ | 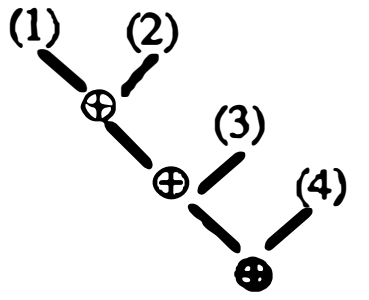  | 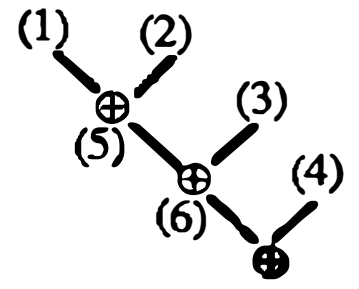  | same |
| $M^{g \to f}$ and $Bel^{\downarrow f}$ | 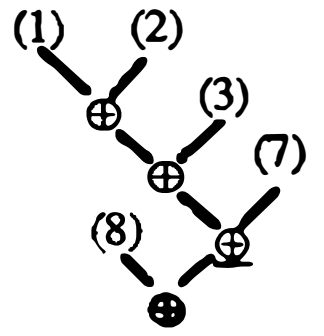  | 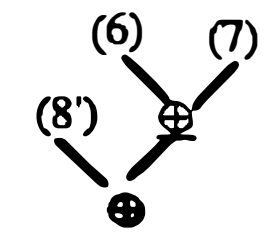  | using (6) instead of re-computing (1)⊕(2)⊕(3) |
| $M^{g \to e}$ and $Bel^{\downarrow e}$ | 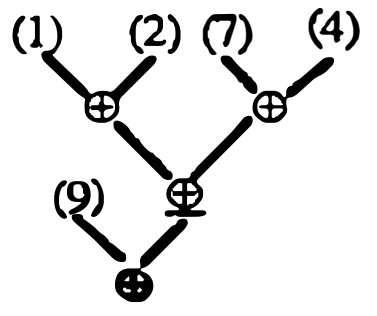  | 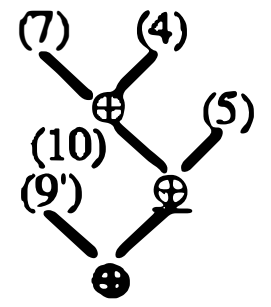  | using (5) instead of re-computing (1)⊕(2) |
| $M^{g \to d}$ and $Bel^{\downarrow d}$ | 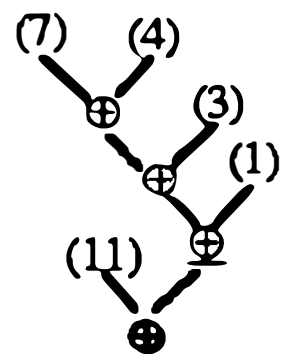  | 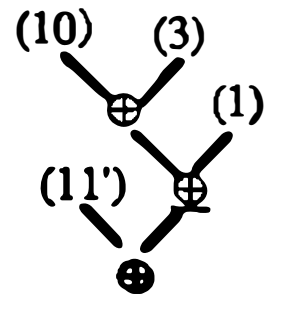  | using (10) instead of re-computing (7)⊕(4) |

The numbers correspond to: (1): $Bel_g$, (2): $M^{d \to g}$, (3): $M^{e \to g}$, (4): $M^{f \to g}$, (5): $\mathcal{Intm}_e$, (6): $\mathcal{Intm}_f$, (7): $M^{h \to g}$, (8): $Bel_f \oplus \{\oplus M^{z \to f} \mid z \in \mathcal{N}_f - \{g\}\}$, (8'): $\mathcal{Cur}_f$, (9): $Bel_e \oplus \{\oplus M^{z \to e} \mid z \in \mathcal{N}_e - \{g\}\}$, (9'): $\mathcal{Cur}_e$, (10): $R_e$, (11): $Bel_d \oplus \{\oplus M^{z \to d} \mid z \in \mathcal{N}_d - \{g\}\}$, (11'): $\mathcal{Cur}_d$.

The following algorithm implements the simultaneous belief functions propagation according to the scheme described above. Note that $\mathcal{Ch}_v$ and $\mathcal{P}_v$ used here always refer to the children and the parent of node v in the same chosen G. In the algorithm, we use "nil" to represent the vacuous belief function. Before propagation, $\mathcal{Cur}_v$ is initiated to $Bel_v$. The process begins by calling **Propagate** $(\mathcal{M}, \mathcal{E})$.

```
                Propagate (M', E')
if M' ≠ ∅ then
   L := L(M', E')
   /* get the leaves of G' = (M',E') */
   if M' = {r} then L := M' end-if
     /* r is the root of G. */
   for i ∈ L    /* for every v in L, do the followings: */
     for k ∈ Ch_i
       /* receive the messages from all its children: */
       Intm_k := Cur_i /* store intermediate result at k*/
       M^{k→i} := ((Cur_k)^{↓(k∩i)})^{↑i}
       /* compute message from k to i */
       Cur_i := Cur_i ⊕ M^{k→i}
       /* combine all the messages from children */
```

end-for
end-for
Propagate($\mathcal{M}$'-L, $\mathcal{E}$'-{(i, j) | i∈ L})
/* delete the leaves, continue propagation in the remaining tree */
for i ∈ L /* for every v in L, do the following: */
if $\mathcal{P}_i$ exists then R := $M^{\mathcal{P}_i \to i}$
/* get the message from the parent*/
else R := nil
/* only root has not parent. */
$Bel^{\downarrow i} := \mathbf{Cur}_i$
/* marginal for the root node is computed*/
end-if
Q := nil
for k ∈ $\mathcal{Ch}'_i$
/* send messages back to every child and compute the marginals for them:*/
R := R ⊕ Q
/* combine some messages from i's neighbours */
$M^{i \to k} := ((\mathbf{Intm}_k \oplus R)^{\downarrow (i \cap k)})^{\uparrow k}$ (4.3)
/* compute message from i to k by using intermediate results*/
$Bel^{\downarrow k} := \mathbf{Cur}_k \oplus M^{i \to k}$
/* marginal for k is computed*/
Q := $M^{j \to i}$
/* get message from k to i for further computation */
end-for
end-for
end-if

Algorithm1 belief function propagation

We can distinguish two parts in the Algorithm 1, separated by the recursive call. In the first part, each node receives the messages from all of its children, and combines them with its own belief function, because the messages are sent starting from the leaf nodes until the root of G is reached, we call this part "propagation-up". In the second part, after each node has received the message from its parent, it sends messages back to its children and computes the marginals for them; as the messages are sent back from the root until the leaves of G are reached, we call this part "propagation-down". Because the leaves of G have no children to receive messages from, the entire propagation can be invoked by calling. **Propagate($\mathcal{M}$-L($\mathcal{M}$,$\mathcal{E}$),$\mathcal{E}$-{(v,$v_j$)| v∈ L($\mathcal{M}$,$\mathcal{E}$)}).**

From Algorithm 1, we find that the number of applications of Dempster's combination at each node is related to the number of its children. Let |S| denote the size of the set S. Generally, in "propagation-up", there are $|\mathcal{Ch}_v|$ combinations at each non-leaf node v; In "propagation-down", there are $2|\mathcal{Ch}_{v_r}|-1$ combinations at the root $v_r$, because $v_r$ has no parent and $3|\mathcal{Ch}_v|-1$ combinations at node v which is neither leaf nor root, thus there are altogether $|\mathcal{Ch}_v|+(2+|\{\mathcal{P}_v\}|)*|\mathcal{Ch}_v|-1$ combinations at non-leaf node v. No combination is needed at leave nodes. In Example 3.1, there are $|\mathcal{Ch}_v|*(1+|\mathcal{N}_v|+\Sigma\{|\mathcal{Ch}_{c_i}| \mid c_i \in \mathcal{Ch}_v, i=1, ..., m, m=|\mathcal{Ch}_v|\})$ combinations at each node.

## 5 UPDATING MESSAGES

Suppose we have already computed the marginals for all the nodes and the Markov tree G is still the same, and we now want to change one or more of the prior belief functions for some reason. Because some of the previous computed intermediate information are stored at each nodes, we can update the marginals for all the nodes without redoing all the work during re-propagation.

For the sake of simplicity, in this section, we use number i to refer to node $v_i$ in the Markov tree, as shown in Fig. 5.1. Node 1 is by convention the root of the tree.

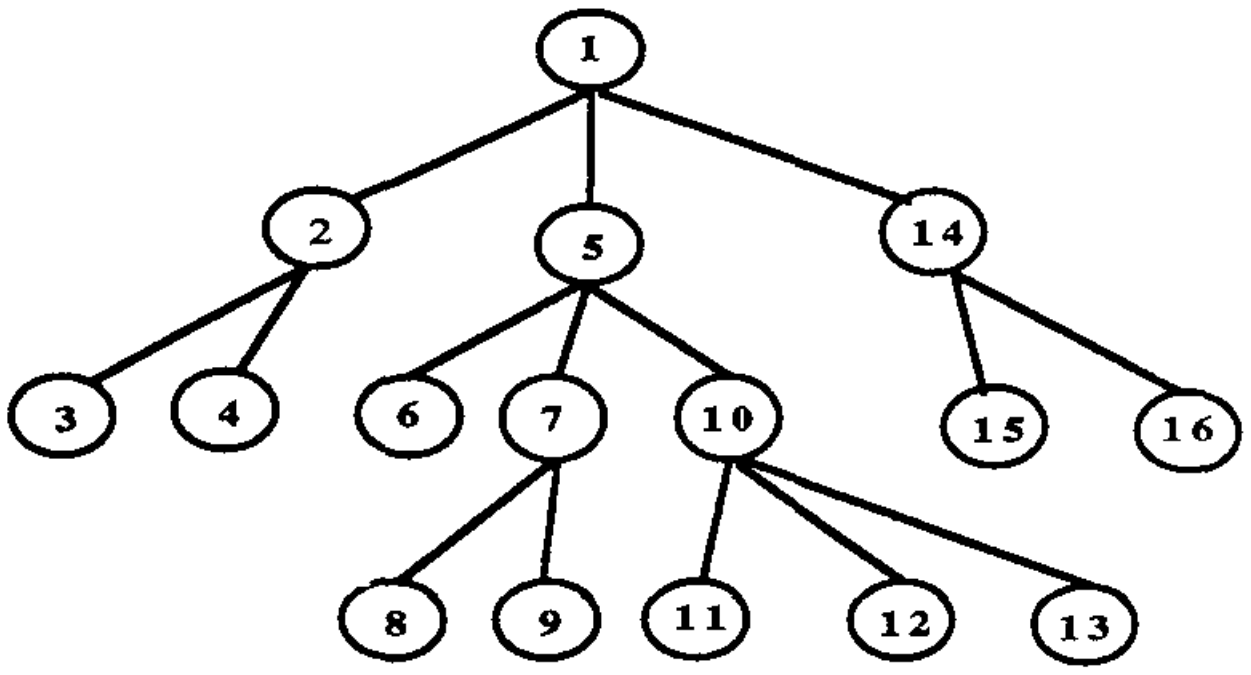

Fig 5.1 a Markov tree representation for a belief function network

Suppose we change one prior belief function, say $Bel_{12}$. According to (3.1), the generic message $M^{i \to j}$ depends on $Bel_i$ and on $M^{k \to i}$ ($k \in \mathcal{N}_i$, $k \neq j$), thus all the messages $M^{i \to \mathcal{P}_i}$ (if $\mathcal{P}_i$ exists) from any i on the path between node 12 and 1, including node1, will be discarded; Moreover, all the messages $M^{\mathcal{P}_j \to j}$ (if $\mathcal{P}_j$ exists) for all j not lying on the path between 12 and 1, will be discarded as well. The remaining half of the messages can be retained. Now suppose we need to compute the marginals for all nodes again. If we have stored all the previous messages, then only the changed messages should be recomputed, while the unchanged ones can be retained.

As an illustration, let's now focus on the computation of $M^{5 \to 1}$, a message which has been discarded by the change in $Bel_{12}$. If there were no $\mathbf{Intm}_{10}$ stored at node 10, we would have to compute $M^{5 \to 1}$ as follows:

$\mathbf{Cur}_5 = Bel_5$
$\mathbf{Cur}_5 = \mathbf{Cur}_5 \oplus M^{6 \to 5} \oplus M^{7 \to 5} \oplus M^{10 \to 5}$
$M^{5 \to 1} = ((\mathbf{Cur}_5)^{\downarrow (5 \cap 1)})^{\uparrow 1}$

i.e. three combinations are needed here.

By using the stored $\mathbf{Intm}_{10}$, we compute $M^{5\to1}$ as:

$\mathbf{Cur}_5 = \mathrm{Bel}_5$
$\mathbf{Cur}_5=\mathbf{Intm}_{10}\oplus M^{10\to5}$ ;as $\mathbf{Intm}_{10}$ is not changed.
$M^{5\to1}=((\mathbf{Cur}_5)^{\downarrow(5\cap1)})^{\uparrow1}$

i.e. just one combination is needed here.

Formally, suppose that one input $\mathrm{Bel}_i$ is changed. According to (3.1), all the messages $M^{k\to\mathbf{P}_k}$ (if $\mathbf{P}_k$ exists) from any k on the path between node i and r (the root), including i, will be discarded; Moreover, all the messages $M^{\mathbf{P}_j\to j}$ (if $\mathbf{P}_j$ exists) for all j not lying on the path between i and r, will be discarded as well. According to (4.1), $\mathbf{Cur}_k$ of node k stores the combination of all the messages sent from the children of k with its own belief function. So only $\mathbf{Cur}_k$ of k lying on the path between i and r, including i and r, will be discarded. According to (4.2), $\mathbf{Intm}_j$ depends on $\mathrm{Bel}_{\mathbf{P}_j}$ and $M^{k\to\mathbf{P}_j}$ ($k\in \mathbf{Lsb}_j$), so for any k on the path between i and r, including i, if $j\in\mathbf{Rsb}_k$, then $\mathbf{Intm}_j$ will be discarded. This point is illustrated in Fig. 5.2 by showing some cases when one prior belief function is changed.

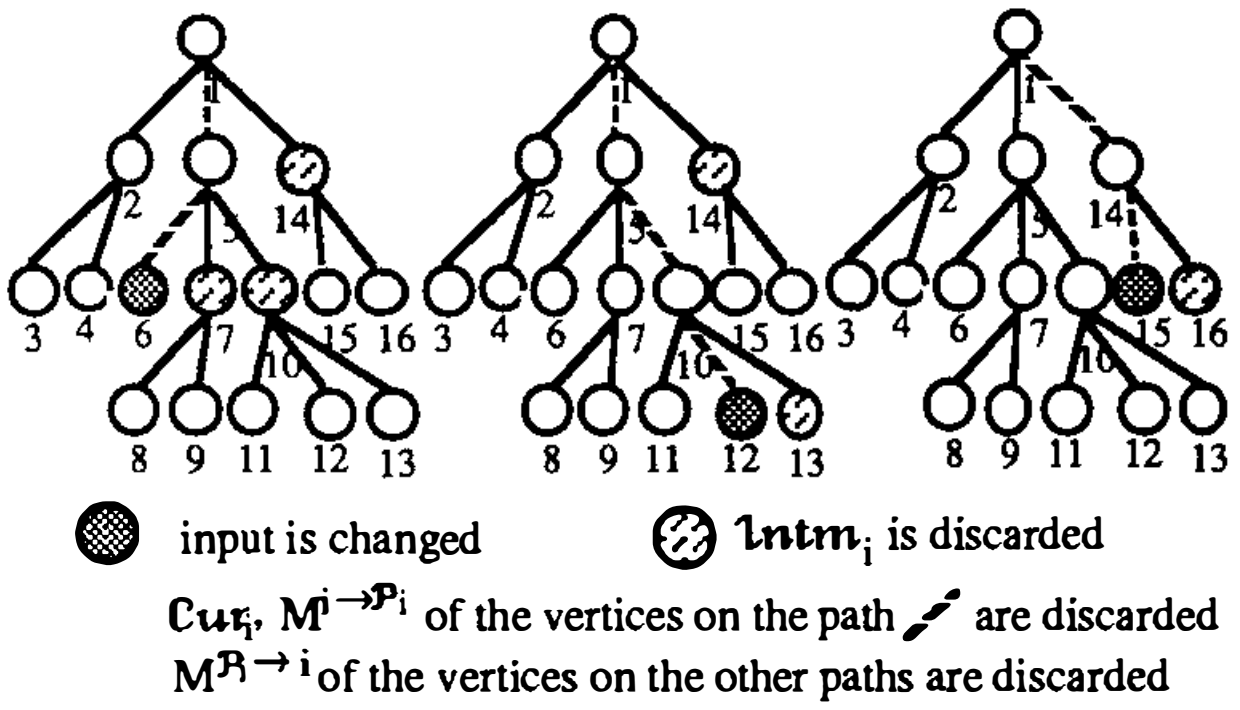

Fig5.2: cases for the changes of the messages

Suppose that for the node i, $\mathbf{Cur}_i$ is not changed, i.e. $\mathrm{Bel}_i$ and all the $M^{k\to i}$ ($k\in\mathbf{Ch}_i$) are not changed. As a consequence, all $\mathbf{Intm}_k$ ($k\in\mathbf{Ch}_i$) are not changed. So whenever $\mathbf{Cur}_i$ is unchanged, we can skip the "propagation-up" part for i during re-propagation. e.g. We want to compute $\mathrm{Bel}^{\downarrow14}$ now. Because $M^{14\to1}$ does not change, it is desirable not to re-compute $\mathrm{Bel}_{14}\oplus M^{15\to14}\oplus M^{16\to14}$ for computing $\mathrm{Bel}^{\downarrow14}$. By using $\mathbf{Cur}_{14}$, we just avoid this computation.

Synthesizing all the cases discussed in these two sections, we give a combined algorithm for both simultaneous propagation and re-propagation. This algorithm is based on the assumption that the structure of the Markov tree is not changed when re-propagating and that $M^{i\to j}$, $\mathbf{Intm}_i$, $\mathbf{Cur}_i$ are only discarded when necessary as explained before. In this Algorithm, we will use $\mathrm{Bel}^{\downarrow i}$ temporarily to compute $\mathbf{Cur}_i$ in the propagation-up part. Initially, for each node i, $\mathrm{Bel}^{\downarrow i}$ is initiated to $\mathrm{Bel}_i$ and for the first child $c_1$ of i, $\mathbf{Intm}_{c_1}$ is initiated to $\mathrm{Bel}_i$. Then the propagation can begin by calling **Propagate($\mathcal{M}$-L($\mathcal{M}$,$\mathcal{E}$), $\mathcal{E}$-{(i,j)|i∈ L($\mathcal{M}$,$\mathcal{E}$)}).**

```
                    Propagate (M', E')
if M' ≠ ∅ then
 L := L(M', E')    /* get the leaves of G' = (M',E') */
 if M'= {r} then L := M' end-if
   /* r is the root of G. */
 for i ∈ L     /* for every v in L, do the followings: */
   if Cur_i does not exist then
   /* if Cur_i exists, the first part is skipped for i */
     Ch := Ch_i    /* otherwise: */
     for j ∈ Ch_i
    /* find the first child i whose message is discarded */
       if M^{j→i} exists then Ch := Ch - {j}
                      else Bel^{↓i} := Intm_j
                           exit loop-for
       end-if
     end-for
     if Ch=∅ then Ch := Ch_i end-if
   /* if all the messages from the children is not changed,
     as Cur_i is discarded, Bel_i is changed. */
     for k ∈ Ch
   /* compute the messages from the rest of children. */
       Intm_k:=Bel^{↓i} /*store intermediate result at k*/
       if M^{k→i} does not exist then
         M^{k→i}:=((Cur_k)^{↓(i∩k)})^{↑i} end-if
   /* compute message from k to i if necessary */
       Bel^{↓i} := Bel^{↓i} ⊕ M^{k→i}
   /* combine all the messages from children */
     end-for
     Cur_i := Bel^{↓i}
   /* store the combination of all the messages from
children of i with the prior belief function of i at node i*/
   end-if
 end-for
 Propagate(M'-L, E'-{(i, j) | i∈L})
    /* delete the leaves, continue propagation in the
remaining tree */
 for i ∈ L(M', E')
   /* for every v in L, do the following: */
     if P_i exists then R := M^{P_i→i}
   /* get the message from the parent */
                     else R := nil
     end-if
     Q := nil
     for k ∈ Ch'_i
   /* send messages back to every child and compute the
marginals for them:*/
       R := R ⊕ Q
   /* combine some messages from i's neighbours */
```

if $M^{i \to k}$ does not exist then
$M^{i \to k} := ((\mathbf{Intm}_k \oplus R)^{\downarrow (i \cap k)})^{\uparrow k}$ end-if
/* compute message from i to k by using intermediate results*/
$Bel^{\downarrow k} := \mathbf{Cur}_k \oplus M^{i \to k}$
/* marginal for k is computed*/
$Q := M^{k \to i}$
/* get message from k to i for further computation */
end-for
end-for
end-if

Algorithm2 combined algorithm for simultaneous propagation and re-propagation

# 6 CONCLUSIONS

We have presented an algorithm (Algorithm 1) for belief function propagation based on the local computation technique proposed by Shafer, Shenoy and Mellouli. The advantage of Algorithm 1 is that it decreases the number of applications of Dempster's combination rule during the propagation, thus reducing the overall complexity, which is one of the most serious problems in implementing Dempster-Shafer theory. Moreover, Algorithm 2 makes full use of the already computed messages and intermediate results for re-propagating when one or more of the prior belief functions is changed. A propagation system* has been implemented in Allegro Common Lisp with Common Windows (by Franz Inc) according to Algorithm 2. It runs on a SUN-3/60 Workstation under SUN Operating System 4.0.3. It has shown that the speed of computation can be greatly increased in comparison with an existing implementation such as MacEvidence (Hsia and Shenoy 1989). Because of the generality of the local computation technique (Saffiotti 1989, Shenoy 1989), this approach is not just specific to belief function propagation, but may be used for any case in which the local computation technique may be applied.

**Acknowledgements**

I am grateful for the support, help and suggestions from Professor Philippe Smets. For the useful discussions and worthy comments on the drafts, I greatly thank Alessandro Saffiotti, Yen-Teh Hsia, Robert Kennes and Elisabeth Umkehrer .This work has been supported by the grant of IRIDIA, Université Libre de Bruxelles.

*The implementation is a property of IRIDIA-Université Libre de Bruxelles, and it is freely available for strictly non-commercial use.